# Semantic 3D Map Change Detection and Update based on Smartphone Visual Positioning System

Max Jwo Lem Lee, Li-Ta Hsu

*Abstract*— Accurate localization and 3D maps are increasingly needed for various artificial intelligence based IoT applications such as augmented reality, intelligent transportation, crowd monitoring, robotics, etc. This article proposes a novel semantic 3D map change detection and update based on a smartphone visual positioning system (VPS) for the outdoor and indoor environments. The proposed method presents an alternate solution to SLAM for map update in terms of efficiency, cost, availability, and map reuse. Building on existing 3D maps of recent years, a system is designed to use artificial intelligence to identify high-level semantics in images for positioning and map change detection. Then, a virtual LIDAR that estimates the depth of objects in the 3D map is used to generate a compact point cloud to update changes in the scene. We present an excellent performance of localization with respect to other state-of-the-art smartphone positioning solutions to accurately update semantic 3D maps. It is shown that the proposed solution can position users within 1.9m, and update objects with an average error of 2.1m.

*Index Terms*—Semantic mapping; Visual positioning system; Dynamic environment

## I. INTRODUCTION

ACCURATE 3D spatial information is crucial for the future artificial intelligence based applications of IoT [1]. This information is stored in a 3D map, which is widely used as prior knowledge for localization, intelligent planning, city services management etc. There are many proposed localization and 3D map generation methods using Simultaneous Localization and Mapping (SLAM) algorithms with cameras and Laser Imaging Detection and Ranging (LIDAR) sensor [2, 3]. However, these large scale high-accuracy maps incur high generation costs and can contain inappropriate information and noise. To consistently recreate a new 3D map using SLAM and LIDAR is expensive and unsustainable [4]. With the rise of smart cities, 3D maps have been developing rapidly and have already become widely available [5, 6]. Therefore, efforts have been devoted to developing accurate 3D map update techniques based on a previous existing 3D map.

Current discussion on outdoor 3D map update is focused on high definition (HD) maps which are designed towards autonomous vehicles around roads [7, 8]. The current 3D map update consists of two main element techniques: change detection and object update [9]. Change detection is one of the major topics in remote sensing, and many methods and application examples have been proposed [10]. A popular method first estimates the position of the vehicle followed by the detection of changed objects in the occupied space with the use of camera images and optical distance sensors. Although the camera image-based methods incur low cost, they cannot accurately measure shapes [11]. 3D reconstruction techniques using stereo-images or multi-view images overcomes this shortcoming, but its performance is sensitive to shooting conditions. Therefore, optical distance sensor-based methods are used for accurate shape measurement. These methods generate dense 3D point cloud maps using LIDAR and detect changed objects by comparing their local densities or shape features [10]. Although they show excellent accuracy, the adoption of professional-grade camera and LIDAR system, coupled with complex image stitching software is nonetheless a costly solution and are normally accessible only to luxury road vehicles.

For a sustainable 3D map update system, it should consider the dynamic human living environment such as the urban and indoor environment. The solution should also be accurate, efficient, and inexpensive. Nowadays, a personal smartphone is equipped with a camera vision sensor. This camera can also be used for localization and as a 3D map update system. The requirement for being easy to deploy and user friendly are also satisfied. The main motivation is that high-performance modern smartphones provide storage, data processing, and cloud-based integration for collaborative 3D mapping that can be easily exploited. This allows 3D map updating coming from multiple smartphones during concurrent or disjoint acquisition session. Currently, there are several studies that make use of smartphone images for localization and 3D map updates. With regard to smartphone localization, Wi-Fi based localization has become extremely popular in the indoor areas and many researchers are focused in this area [12]. However, the use of Wi-Fi is still very challenging, suffering tens of meters even in strong signal conditions [13], where the calibration of Wi-Fi fingerprinting database and the density of Wi-Fi beacons in indoor environment pose a lot of challenges. In the context of outdoor pedestrian localization, the application of global navigation satellite system (GNSS) is the key technology to provide accurate positioning/timing service in open field environments. Unfortunately, its positioning performance in urban areas still has a lot of potential to improve due to signal blockages and reflections caused by tall buildings and dense foliage [14]. In such environments, most signals are non-line-of-sight (NLOS) which can severely degrade the localization accuracy [15]. An

This work is supported by PolyU RISUD on the project – BBWK "Resilient Urban PNT Infrastructure to Support Safety of UAV Remote Sensing in Urban Region."

M.J.L. Lee and L-T. Hsu are with Interdisciplinary Division of Aeronautical and Aviation Engineering, The Hong Kong Polytechnic University (PolyU). L-T. Hsu is also with Research Institute for Sustainable Urban Development (RISUD). S. Lee is with Hong Kong University of Science and Technology School of Engineering. Corresponding author: Li-Ta Hsu (e-mail: lt.hsu@polyu.edu.hk).

idea called GNSS shadow matching was proposed to improve urban positioning [16]. It first classifies the received satellite visibility by the received signal strength and then scans the predicted satellite visibility in the vicinity of the ground truth position. Position is then estimated by matching the satellite visibilities. Another method uses ray-tracing–based 3DMA GNSS algorithms that cooperate with pseudorange have been proposed [17]. The integration of the shadow matching and range-based 3DMA GNSS are proposed in [18]. Where the performance of this approach in multipath mitigation and NLOS exclusion depends on the accuracy of the 3D building models [19]. With regard to 3D map update, an area of study focuses on depth estimation using monocular SLAM. The concept is to use visual odometry and structure from motion (SfM) on captured images [20]. These applications identify low level features in current frame with those extracted from previous frames, to map the 3D structure of a scene. These algorithms can produce a point cloud-based 3D map similar to LIDAR without the need of expensive equipment. Although a suitable method for a 3D reconstruction of an environment from 2D images, it does not reuse the existing 3D map, and requires a large amount of processing power. In addition, the approach suffers from the need of successive images to calculate the accurate depth in images. These successive images must be taken in short intervals and not vary significantly to avoid dynamic interference. The generated point cloud may also be spare with limited accuracy to properly capture the depth of objects in images. Therefore, another area of study focuses on updating low-level features only. Google's recently developed feature-based visual positioning system (VPS) identifies edges within the smartphone image and matches with edges captured from pre-surveyed images in their map database for localization [21]. These pose-tagged edges are stored in a searchable index that can be updated overtime using user images. However, the solution only updates low level features, which makes it useful for localization but lacks contextual information that is often highly valued in a 3D map.

Thus, detection solely on the low-level features may not be enough for precise positioning and 3D map update. Standing at the point of view of how a pedestrian navigate him/herself, in addition to the identification of features, we, human beings, also locate based on the materials and objects that consists of different semantic information, and each semantic has a class of its own. We also remember our environment and recognize any changes in the environment. Therefore, in contrast to the existing solutions, this paper focus on a novel semantic 3D map change detection and update solution that overcomes the aforementioned difficulties for the urban and indoor environments. In our previous paper, we proposed a novel semantic VPS using different types of classes that are widely seen and continuously distributed in the urban scene [22]. This has achieved near 2m state-of-the-art positioning accuracy. This positioning error can be reduced if the encountered inconsistencies in the 3D map are updated to reflect reality. Inspired from existing 3D map update solution, our proposed novel solution is the semantic 3D map detection and update by utilizing the combination of semantic VPS, virtual LIDAR, and descriptors. The proposed method is a continuation of the semantic VPS research. In addition to overcoming the aforementioned weaknesses of existing 3D map update methods, it also offers several major advantages:

- Firstly, using the semantic VPS, we can take advantage of existing 3D maps as visual aids for precise self-localization, improving conventional positioning inaccuracies which are common in the urban and indoor environments.
- Secondly, high level semantics are used as the medium to detect any change in class in the image, a research domain that has been pioneered.
- Thirdly, the use of virtual LIDAR in the 3D map can provide an approximate dimension and pose of the changed class detected in the smartphone image in the world coordinate system.
- Fourthly, descriptors are used to describe the dimension and pose of the class in the 3D map, making the descriptors easy to add, replace and remove.
- Fifthly, the descriptors used to update the 3D map retains the important semantic contextual information which are often valued in a 3D map.
- Lastly, the consideration of dynamic objects (pedestrians, etc.) as descriptors, such that it can be extended to a dynamic 3D map.

The proposed solution in this research uses the improved VPS pose estimation for the semantic 3D map update. We believe this research demonstrates a very good solution to provide seamless localization and 3D map update for many futuristic artificial intelligences based IoT applications, such as augmented or virtual reality. The remainder of the paper are organized as follows. Sect 2 explains the overview of the proposed semantic 3D map change detection and update approach. Sect. 3 describes the semantic VPS, change detection and descriptor update in detail. Sect. 4 describes the experimentation process and the results compared to ground truth. Sect. 5 contains the concluding remarks and future work.

## II. OVERVIEW OF THE PROPOSED METHOD

An overview of the proposed semantic 3D map update is shown in Fig. 1. The method is divided in two main stages: an offline process, and an online process. In the offline process, a textured 3D map is manually segmented into different colors based on different classes (Sect. 3.1). This segmented 3D map will form the basis of the 3D map to be updated. In the online process, the user captures an image with their smartphone, with the initial pose (3D position, 3D rotation) and camera intrinsic parameters recorded (Sect. 3.2). The smartphone image is then manually segmented between the different types of classes identified in the image (Sect. 3.3). The segmented smartphone image, initial pose, intrinsic parameters, and the segmented 3D map is used to estimate an improved VPS pose solution (Sect. 3.4). Then, the VPS pose and intrinsic parameters are used to generate an image (Sect. 3.5) with a virtual camera inside the segmented 3D map. A local change detector algorithm detects the difference in class between the generated image and the smartphone image; the changed class in the smartphone image is identified (Sect. 3.6). The VPS pose and intrinsic parameters are also used to generate a point cloud (Sect. 3.7) with a and virtual LIDAR inside the segmented 3D map. The point cloud is projected to the image coordinate system, where each pixel in the image has a corresponding 3D position in local coordinates (Sect. 3.7). The local coordinates of the point cloud are transformed to the world coordinates (Sect. 3.8). A descriptor is assigned to describe the dimension, position, and orientation of the changed class, calculated based on the pixels that corresponds to a 3D position in world coordinates (Sect. 3.9). The descriptor is then placed into the updated semantic 3D map (Sect. 3.9).

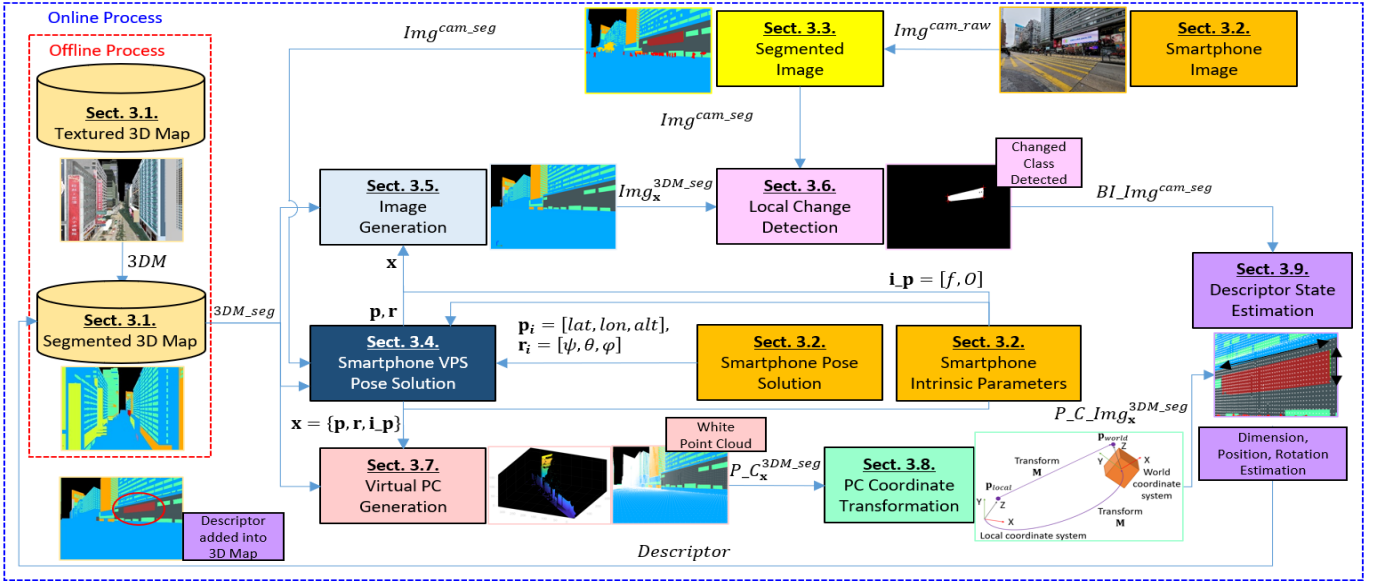

Fig 1. Flowchart of the semantic 3D map change detection and update based on smartphone VPS.

## III. PROPOSED METHOD IN DETAIL

### 3.1 Segmented 3D Map

The outdoor 3D map used in this research is provided by the *Surveying and Mapping Office, Lands Department, Hong Kong* [23]. It consists of only buildings and infrastructures. The indoor 3D map is provided by the *Hong Kong Polytechnic University* and consists of the indoor environment of a library. The 3D maps consist of level of detail 2, stored as *Autodesk 3ds Max* (3ds) format. Each object inside the 3D maps has its own corresponding 2D vector map in JPG format that provides textural information of the object. The object vector maps were manually segmented with the Image Labeler application, which is part of the *Computer Vision Toolbox, MATLAB* [24]. Each pixel in the texture image is assigned a color for the class it represents, which can then be used to simulate a segmented 3D map as shown in Fig. 1. Each material class has their own respective RGB color: Stone (blue), Glass (green), Metal (orange), Banner (brown). To test the feasibility of the proposed method, these material classes in the 3D map will be updated based on the changed classes detected in the smartphone image. In addition, dynamic object classes such as Pedestrian (red) and Chair (pink) are also tested as changed classes to update the 3D maps. Lastly, Sky (black), Foliage (yellow), Others (light blue) will be detected but not be used to update the 3D map due to its large variance in dimension.

The 3D visualization software to automatically update the semantic 3D map based on the detected changed classes is performed in *Unreal Engine, Epic Games* [25]. The semantic 3D map uses the 3D Cartesian meter coordinate system on a plane to determine the positioning coordinates. Therefore, it was necessary to convert the measured geographic coordinate system in (latitude and longitude) back to the 3D Cartesian coordinate. Thus, we transform between the WGS84 Geographic coordinates and Hong Kong 1980 Grid coordinates using the equations described by the *Surveying and Mapping Office, Lands Department, Hong Kong* [23].

### 3.2 Smartphone image acquisition and format

Since the semantic 3D map change detection and update is based upon the detected classes within the image, the localization is likely to perform well when there is a richer and more diverse class distribution in the image. Therefore, the widest available angle lens is the preferred choice as it is more suitable to capture more classes of the surrounding scene in the image. A conventional smartphone camera with a 120° diagonal field of view, 4:3 aspect ratio, resolution of [960,720] pixels is used to capture images shown in Fig. 1. The initial position and rotation recorded by the smartphone when an image is taken is denoted as:

$$\mathbf{p}_i = [lat, lon, alt]$$
$$\mathbf{r}_i = [\psi, \theta, \varphi] \quad (1)$$

Where $\mathbf{p}_i$ is the three-dimensional initial position and $\mathbf{r}_i$ is the three-dimension initial rotation. The camera intrinsic parameters ($\mathbf{i\_p}$) including the focal length and the optical center can be denoted as:

$$\mathbf{i\_p} = [f, O]$$
$$\mathbf{x}_i = \{\mathbf{p}_i, \mathbf{r}_i, \mathbf{i\_p}\} \quad (2)$$

$\mathbf{x}_i$ is the state that defines the initial pose which holds the initial position, initial rotation and intrinsic parameters.

### 3.3 Segmented Image

The captured smartphone images were segmented manually with the Image Labeler application in MATLAB to output the ideal segmented smartphone image. In the future, however, we plan to utilize use an artificial intelligence deep learning neural network to identify the class automatically.

$$Img^{cam\_seg} = SEG(Img^{cam\_raw}) \qquad (3)$$

Where $SEG$ is the function to segment the raw smartphone image. The segmented smartphone image is denoted as $Img^{cam\_seg}$. The format of the segmented smartphone image can be described as:

$$Img^{cam\_seg} = SI(\mathbf{u}, \mathbf{v})$$
$$SI \in \{\text{Stone (0), Glass (1), Metal (2), Banner (3),}$$
$$\text{Pedestrian (4), Chair (5), Sky (6), Foliage (7),} \qquad (4)$$
$$\text{Others(8)}\}$$

Where $\mathbf{u}, \mathbf{v}$ are the 2D pixel coordinates of the pixel inside the image. $SI$ is the function that assigns each pixel an indexed number to represent a class. Fig. 1 shows an example of a segmented smartphone image.

### 3.4 Smartphone VPS Pose Solution

A semantic VPS method is used to estimate the improved pose solution. The initial pose estimated by the smartphone GNSS receiver, and IMU sensors were used to generate candidates (hypothesized poses) that are spread across a searching grid around the initial pose. The segmented smartphone image is matched with the candidate images using multiple metrics to calculate the similarity scores. The scores of each method are combined to calculate the likelihood of each candidate. Then, the VPS pose is determined by the candidate with the maximum likelihood among all the candidates. The details of the proposed method are described in [22]. The improved pose is be denoted as:

$$\mathbf{x} = VPS(Img^{cam\_seg}, 3DM\_seg, \mathbf{x}_i)$$
$$\mathbf{x} = \{\mathbf{p}, \mathbf{r}, \mathbf{i\_p}\} \qquad (5)$$

Where $3DM\_seg$ is the segmented 3D map, and $VPS$ is the function to calculate the improved pose solution based on the segmented smartphone image, segmented 3D map and the initial state. $\mathbf{x}$ is the state that defines the pose which holds the improved VPS pose estimation.

### 3.5 Image generation

The improved VPS pose is then used to generate an image from the segmented 3D map using a virtual camera. The generated image is compared against the segmented smartphone image to calculate the difference in class.

$$Img_{\mathbf{x}}^{3DM\_seg} = RL\_P(3DM\_seg, \mathbf{x}) \qquad (6)$$

Where $RL\_P$ is the function to capture the image in the 3D map using a virtual camera based on the VPS pose. The generated image is denoted as $Img_{\mathbf{x}}^{3DM\_seg}$. Fig. 1 shows an example of an image generated from the segmented 3D map based on a VPS state. The format of the images can be described as:

$$Img_{\mathbf{x}}^{3DM\_seg} = SI(\mathbf{u_x}, \mathbf{v_x}) \qquad (7)$$

Where $\mathbf{u_x}, \mathbf{v_x}$ are the 2D pixel coordinates of the pixel inside the image generated based on the VPS state $\mathbf{x}$.

### 3.6 Local Change Detection

The local change algorithm calculates the difference of each class between the segmented smartphone image and the VPS generated image. The target function is to find the region (in pixels) of the changed class in the segmented smartphone image. An absolute difference algorithm was used to detect the change. Described in Eq. 4, each pixel coordinate in a segmented image has an indexed number that represents a class. The indexed number (representing the class) of each pixel of the images are subtracted. The changed pixel region is determined by the pixel coordinate where the subtracted indexed number representing the class is not zero.

$$BI\_Img^{cam\_seg}$$
$$= \begin{cases} 1 & \left|Img^{cam\_seg}(\mathbf{u}, \mathbf{v}) - Img_{\mathbf{x}}^{3DM\_seg}(\mathbf{u_x}, \mathbf{v_x})\right| > 0 \\ 0 & \text{otherwise} \end{cases} \qquad (8)$$
$$BI\_Img^{cam\_seg} = BI(\mathbf{u}, \mathbf{v})$$
$$BI \in \{\text{no change (0), change (1)}\}$$

$BI$ is the function that assigns each pixel a binary number to represent a changed class in the segmented smartphone image. The change detector image is denoted as $BI\_Img^{cam\_seg}$. A filtering algorithm is then performed to remove detected changed regions in the image with less than 5%-pixel region [26]. Where 100% is the total number of pixels in the image. The 5% threshold is calibrated using the experiments presented in this research. There are two purposes regarding to the filtering algorithm. Firstly, to remove noise in the detected regions that does not constitute as a change. Secondly, to remove actual changes that are too small which cannot be accurately measured using the proposed solution. Once the change detector image identifies the pixel coordinates of the changed class, the class itself can be identified in the index of the corresponding pixel coordinates of the segmented smartphone image. Hence, the changed area and the class occupied by the changed area in the segmented smartphone image can be obtained.

### 3.7 Point cloud Generation

Point cloud were generated from the segmented 3D map using a virtual LIDAR to estimate the position of the classes captured in the smartphone image, using *Computer Vision Toolbox, MATLAB* [24]. An assumption was made that the 3D map resemble reality enough to provide an approximate distance between a point and an object in the 3D map. Taking advantage of this distance, a virtual LIDAR is then placed in the 3D map at the VPS pose.

$$P\_C_{\mathbf{x}}^{3DM\_seg} = P\_C\_GEN(3DM\_seg, \mathbf{x}) \qquad (9)$$

Where $P\_C\_GEN$ is the function to capture the point cloud in the 3D map using a virtual LIDAR based on the VPS state (VPS pose and the intrinsic parameters). The point cloud generated in world coordinates are denoted as $P\_C_{\mathbf{x}}^{3DM\_seg}$. Fig. 1 shows an example of a point cloud generated from the segmented 3D map. The virtual LIDAR has a resolution of 0.1m and captures a point cloud of the objects it detected. The point cloud is in the local 3D coordinate frame relative to the virtual LIDAR and stored as a list of 3D points. We first projected the point cloud to the image coordinate system, so that we can assign each pixel a 3D position in the smartphone segmented image. The coordinate system for camera projection is shown in Fig 2.

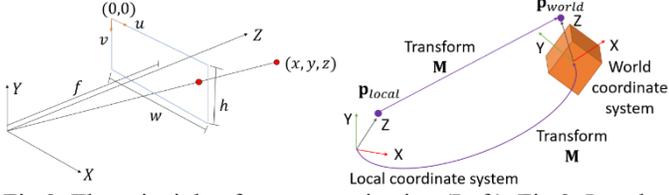

Fig 2. The principle of camera projection (Left). Fig 3. Local to world coordinate transformation (Right).

The image plane is located along the optic axis by a focal length of $f$, as shown in Fig 2. The principle of a pinhole camera projection is used for relating a list of LIDAR 3D points to a 2D image. Z is the optic axis. The X Y plane parallels the focal length plane. We complete the perspective projection via the triangle rule. Therefore, we can get the correspondence between the local 2D image coordinates and local 3D coordinates.

$$u = f\frac{x}{z} \qquad v = f\frac{y}{z} \qquad (10)$$

Where $(x, y, z)$ is a point position in the 3D local coordinate, and $(u, v)$ is a pixel position in the 2D pixel coordinate, $f$ is the focal length. In this way it is possible to estimate the 3D position of a pixel in the local 3D coordinates from its projection in the image in pixels. The process can be described as:

$$P\_C\_L\_Img_x^{3DM\_seg} = C\_T(P\_C_x^{3DM\_seg}) \qquad (11)$$

Where $C\_T$ is the function to relate the local 3D coordinates of the point cloud to the image coordinate system based on Eq. 10. The projected local point cloud on the image is denoted as $P\_C\_L\_Img_x^{3DM\_seg}$. Fig. 1 shows an example of a point cloud image generated from the segmented 3D map. The format of the point cloud image can be described as:

$$P\_C\_L\_Img_x^{3DM\_seg} = P\_I(\mathbf{u_x}, \mathbf{v_x})$$
$$P\_C\_L\_Img_x^{3DM\_seg} \in \mathbf{p}_{local} \qquad (12)$$

Where $\mathbf{u_x}, \mathbf{v_x}$ are the 2D pixel coordinates of the pixel inside the image generated based on the VPS state $\mathbf{x}$. $P\_I$ is the function that assigns each pixel a 3D local position.

### 3.8 Point Cloud Coordinate Transformation

Finally, we transform the point cloud from the local image coordinate to the world image coordinate using a 3D rigid geometric transformation, as shown in Fig 3. The 3D world coordinates are related to the 3D local coordinates with respect to the VPS pose. The rigid transformation moves each point while preserving the distance between points in the cloud. The 3D rigid transformation is parameterized by a 4x4 transformation matrix in homogeneous coordinates.

$$\mathbf{p}_{world} = \mathbf{p}_{local} \cdot \mathbf{M}$$
$$M \in \{\mathbf{p}, \mathbf{r}\} \qquad (13)$$

Where $\mathbf{M}$ is the transformation matrix corresponding to the VPS pose. Using the transformation matrix, the local-to-world coordinate system can be calculated.

$$P\_C\_W\_Img_x^{3DM\_seg} = P\_I(\mathbf{u_x}, \mathbf{v_x})$$
$$P\_C\_W\_Img_x^{3DM\_seg} \in \mathbf{p}_{world} \qquad (14)$$

$P\_I$ is the function that assigns each pixel a 3D world position. After this process, each pixel in the smartphone segmented image can correspond to a 3D position in world coordinates.

### 3.9 Descriptor State Estimation

The descriptor state estimation is used to update the classes in the 3D map based on the classes detected in the change detector image. These classes are split into two categories. The material class (stone, glass, metal, banner) are updated as a 2D dimensional descriptor that are placed on the building surfaces in the 3D map. The dynamic class (pedestrian, chair) are updated as a descriptor with a pre-defined 3D dimension that are placed above the terrain in the 3D map. The point cloud image is first overlayed onto the smartphone segmented image, hence each pixel from the smartphone segmented image can correlate to a 3D position as shown in Fig 1. To reduce complexity, in this research, we only updated rectangular materials. Therefore, four corners were used to detect the shape of the changed material class. The corner detection is based on an efficient algorithm designed for convex polygons proposed in [27]. The 3D positions of the pixel corners were used to estimate the 2D dimension, position, and rotation of the descriptor.

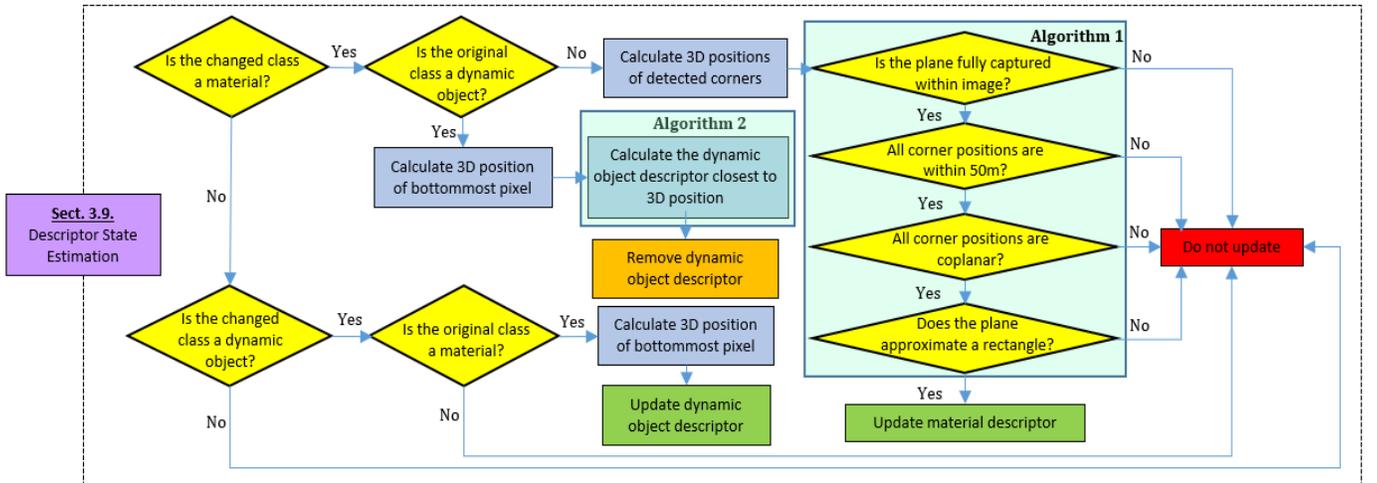

Fig 4. Flowchart of the descriptor update algorithm.

$$M\_Descriptor = M\_STATE\_EST(BI\_Img^{cam\_seg}, P\_C\_Img_x^{3DM\_seg}) \quad (15)$$
$M_{Descriptor} \in \{\text{Material Class, 2D dimension, 3D position, 3D rotation}\}$

Where $M\_STATE\_EST$ is the function that calculates the 2D dimension, position and rotation of a material class. The material descriptor is denoted as $M\_Descriptor$. As shown in the Fig. 4, four requirements need to be fulfilled to update the material descriptor. Firstly, the material object must be captured entirely in the change detection image. Therefore, it is more likely to estimate an accurate dimension of the material descriptor as it does not extend outside of the image. Secondly, the 3D points must be within 50m to the VPS position; as the point cloud accuracy decrease rapidly after 50m [28]. Thirdly, the 3D points of the four corners detected on the change detector image must be coplanar. Updating materials that are not 2D are avoided, as it will significantly increase the positional and dimensional error of the material descriptor. Lastly, the shape must approximate a rectangle. However, it is possible to update different shapes simply by increasing corner detection. For these reasons, if any of the requirements are not fulfilled, not updating the 3D semantic map will be the more appropriate choice. Algorithm 1 provides the steps for the four requirements.

**Algorithm 1** Determining whether to update the changed material class.

| | |
|---|---|
| **Input** | Change detector image, $BI\_Img^{cam\_seg}$ |
| | Detected four corners in world coordinate, $\mathbf{a}(x_1, y_1, z_1), \mathbf{b}(x_2, y_2, z_2), \mathbf{c}(x_3, y_3, z_3), \mathbf{d}(x_4, y_4, z_4)$ |
| | VPS estimated position, $\mathbf{p}(x, y, z)$ |
| **Output** | Material update decision |

1   **for** each pixel in image border, $(\mathbf{u}_b, \mathbf{v}_b)$
2     **if** $BI\_Img^{cam\_seg}(\mathbf{u}_b, \mathbf{v}_b) = 0$
3       Changed class is fully captured
4     **else** Do not update
5     **end if**
6   **end**
7   Calculate distance to first corner,
    $d_1 = |\mathbf{pa}| = \sqrt{(x_1 - x)^2 + (y_1 - y)^2 + (z_1 - z)^2}$
8   Calculate distance to second, third and fourth corners, $d_2 = |\mathbf{pb}|, d_3 = |\mathbf{pc}|$ and $d_4 = |\mathbf{pd}|$
9   **if** $d_1 \& d_2 \& d_3 \& d_4 \leq 50$
10     Corners are within 50m
11   **else** Do not update
12   **end if**
13   Calculate first vector,
    $\vec{ab} = (x_2 - x_1)\mathrm{i} + (y_2 - y_1)\mathrm{j} + (z_2 - z_1)\mathrm{k}$
14   Calculate second and third vector, $\vec{ac}$ and $\vec{ad}$
15   **if** $\vec{ab} \cdot (\vec{ac} \times \vec{ad}) \cong 0$
16     Corners are coplanar
17   **else** Do not update
18   **end if**
19   **if** $\vec{ab} \cong \vec{dc}$ & $\vec{ad} \cdot \vec{ab} \cong 0$
20     Corners are rectangular
21   **else** Do not update
22   **end if**
23   **Update**

To remove a dynamic object descriptor in the semantic 3D map as shown in Fig 4, the absolute distance is calculated between the detected 3D position of the removed dynamic object, and the existing descriptors of the same class in the semantic 3D map. The descriptor with the smallest distance is then removed in the semantic 3D map. Algorithm 2 depicts this process.

**Algorithm 2** Remove the dynamic object descriptor closest to 3D point.

| | |
|---|---|
| **Input** | Detected 3D position of removed dynamic object in world coordinate, $\mathbf{e}(x_5, y_5, z_5)$ |
| | Class of the removed dynamic object, $c$ |
| **Output** | Chosen dynamic object descriptor to remove |

1   **for** each descriptor with the corresponding class in the 3D map, $Descriptor_c \in [\mathbf{p_1}, \mathbf{p_2} \ldots \mathbf{p_x}]$
2     $d = |\mathbf{p_x e}|$
3     update $descriptor\_distance(\mathbf{p}_x) = d$
4   **end**
5   Remove descriptor where $\arg\min_d descriptor\_distance$

For dynamic classes, these classes are usually not in contact with building walls, therefore we cannot estimate dynamic class position based on the building position. Instead, dynamic classes are constantly in contact with the ground due to gravity. We therefore take the position of the bottommost pixel of the dynamic classes detected in the change detection image to estimate the position.

$$D\_Descriptor = D\_STATE\_EST(BI\_Img^{cam\_seg}, P\_C\_Img_x^{3DM\_seg}) \quad (16)$$
$D\_Descriptor \in \{\text{Dynamic Class, 3D dimension, 3D position}\}$

Where $D\_STATE\_EST$ is the function that calculates the position of a dynamic class. The 3D dimension is pre-defined for the dynamic class. Based on the descriptors, an API is used to automatically place the classes into the 3D map.

IV. EXPERIMENTAL RESULTS

*4.1 Image and test location setting*

In this study, the experimental locations were selected within the Tsim Sha Tsui and Hung Hom area of Hong Kong, as shown in Table I. The first three experiments were conducted in the outdoor deep urban environment surrounded by tall buildings in a vibrant human setting where GNSS signals are heavily reflected. Material change on building surfaces are also prominent. The last three experiments were selected in the indoor library environment where GNSS signals are blocked. In both environments, we tested the semantic VPS accuracy and semantic map update of common static and dynamic objects. Three images were taken at each environment using a generic smartphone camera (Samsung Galaxy Note 20 Ultra 5G smartphone with the ultra-wide-lens 13mm 12-MP f/2.2) and a tripod. The ground truth positions of the changed objects were determined based on real life measurements of the objects in relation to identifiable landmarks available on Google Earth, such as a labelled corner on the ground. For each image taken, the ground truth poses were also recorded manually based on Google Earth and nearby identifiable landmarks. Based on the experience of previous researches, the ground truth uncertainty of latitude and longitude is ±1m and yaw is ±2° [22]. The *XPRO geared head, Manfrotto*, is used to measure the altitude with a ±0.1m uncertainty and pitch and roll with ±1° uncertainty, respectively.

TABLE I. Locations and images tested with the proposed semantic VPS.

| Experimental Images | | | | | |
|---|---|---|---|---|---|
| Outdoor Environment | | | Indoor Environment | | |
| 1.1 | 1.2 | 1.3 | 2.1 | 2.2 | 2.3 |
| 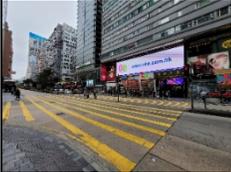 | 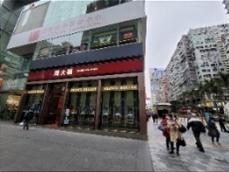 | 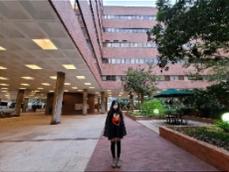 | 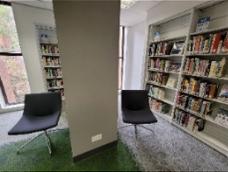 | 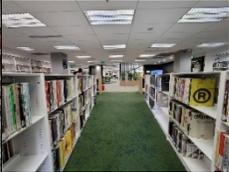 | 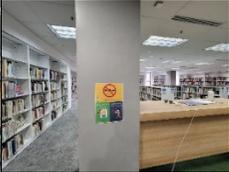 |

## 4.2 Semantic VPS results

The initial rough estimation of the pose is calculated by the smartphone GNSS receiver and IMU sensor when capturing an image with the smartphone. The candidate latitudes and longitudes are distributed around the initial position in a 40m radius with 1m resolution. The candidate yaw is distributed in a 40° angle around the initial yaw with 1° separation. The candidate pitch, roll, and altitude remains the same as measured by the smartphone due to its already high accuracy as demonstrated in [22]. The positioning quality of the proposed method was analyzed based on the ideal smartphone image segmentation. The experimental results were then post-processed and compared to the ground truth, NMEA, and 3DMA.
1. Ground Truth.
2. Proposed semantic VPS [22].
3. NMEA: Low-cost GNSS solution by Galaxy S20 Ultra, Broadcom BCM47755.
4. 3DMA: Integrated solution by 3DMA GNSS algorithm on shadow matching, skymask 3DMA and likelihood based ranging GNSS [29].

**Loc. 1.1** is in a common urban environment with high rise buildings. The results show the positioning accuracy of the semantic VPS improves the positioning accuracy to around two-meter level accuracy as shown in Table II. The semantic VPS is accurate due to the diverse classes available for semantic matching, outperforming 4.59m positioning error of 3DMA. The 10.22m positioning error of NMEA were likely because of the diffraction of GNSS signals passing under high-rise buildings. As shown in Table III, the candidates are higher in similarity (red) as they get closer to the ground truth, the semantic VPS (yellow dot) is very close to the ground truth (white star), achieving 1.96m accuracy. In **Loc. 1.2**, there is a significantly larger area of red candidates surrounding the chosen candidate. This phenomenon can be explained by a condition required for accurate positioning. Ideally there should be no discrepancies between the smartphone image and the candidate image at ground truth. However, as shown in Table V, a large section of the building has changed from stone to glass in the smartphone image. It is likely due to a large difference in class between the smartphone image and candidate image at ground truth, that has caused a high uncertainty in all candidates, thus the candidates share a common similarity and color. This large difference in class will directly decrease the positioning certainty and accuracy. It should be noted that this is still a 3x positioning improvement compared to 3DMA. Although **Loc. 1.1** also has difference in class, the changed class does not occupy a significant area of the smartphone image, hence, the semantic VPS was able to estimate an accurate position. **Loc. 1.3** is surrounded by buildings of the same shape, size, and material. Therefore, it is a very challenging environment for the proposed method as the candidate images share a common class distribution. Nonetheless, a 2.04-meter level accuracy suggests that the semantic VPS can be used as a positioning method in foliage dense environments, outclassing 3DMA of 6.03m.

**Loc. 2** is an indoor library environment where GNSS signals are blocked by walls of the building. The results show the positioning of the proposed method in the indoor environment to be around 1.36m accuracy. The proposed method in **Loc 2.1** has the highest VPS positioning error in the indoor environment of 2.66m. As shown in Table IV, there are only few distinctive classes for the VPS to match with in the smartphone segmented image. This uncertainty has led to a larger area of red candidates as shown in the heat map. It is important to note that even with few distinctive classes, the semantic VPS still improves upon the conventional NMEA solution. **Loc 2.2** demonstrates the strength of the semantic VPS in the presence of more distinctive classes. The shelfs have provided an excellent source of semantic information, with only an error of 0.98m, the VPS achieves a meter-level accuracy. As shown in the heatmap, the number of red candidates is small and close to the ground truth. **Loc 2.3** has achieved the smallest VPS error of 0.45m out of all the experiments. It is likely that having more classes to match with, have aided the positioning accuracy.

In an environment with semantics, the classes can also be used to estimate the heading. The VPS succeeds at predicting the heading accurately within an average of 2 degrees. Hence, the proposed method can be considered an accurate approach to estimate the heading of the user in the urban and indoor environment, outperforming NMEA of 15 degrees.

TABLE II. Positioning and heading performance comparison of the semantic VPS and other positioning methods.

| Loc. | Position and Heading Error from Ground Truth. | | | | | Map Update Error from Ground Truth. | |
|---|---|---|---|---|---|---|---|
| | Semantic VPS | | NMEA | | 3DMA GNSS | Semantic Map Update | |
| | Position Unit: meter. | Heading Unit: degrees. | Position Unit: meter. | Heading Unit: degrees. | Position Unit: meter. | Position Unit: meter. | Dimension Unit: %. |
| 1.1 | 1.96 | 3 | 10.22 | -34 | 4.59 | 1.88 | -44 |
| 1.2 | 3.11 | 4 | 28.95 | 26 | 10.07 | - | - |
| 1.3 | 2.04 | -2 | 17.39 | -1 | 6.03 | 1.57 | - |
| 2.1 | 2.66 | -1 | 5.63 | 5 | - | 2.92 | - |
| 2.2 | 0.98 | 1 | 8.98 | 17 | - | - | - |
| 2.3 | 0.45 | 2 | 9.27 | 5 | - | - | - |
| All Avg. | **1.87** | **2** | **13.41** | **15** | **6.90** | **2.12** | **-44** |

TABLE III. Heatmap on the likelihood of candidates based on the semantic VPS.

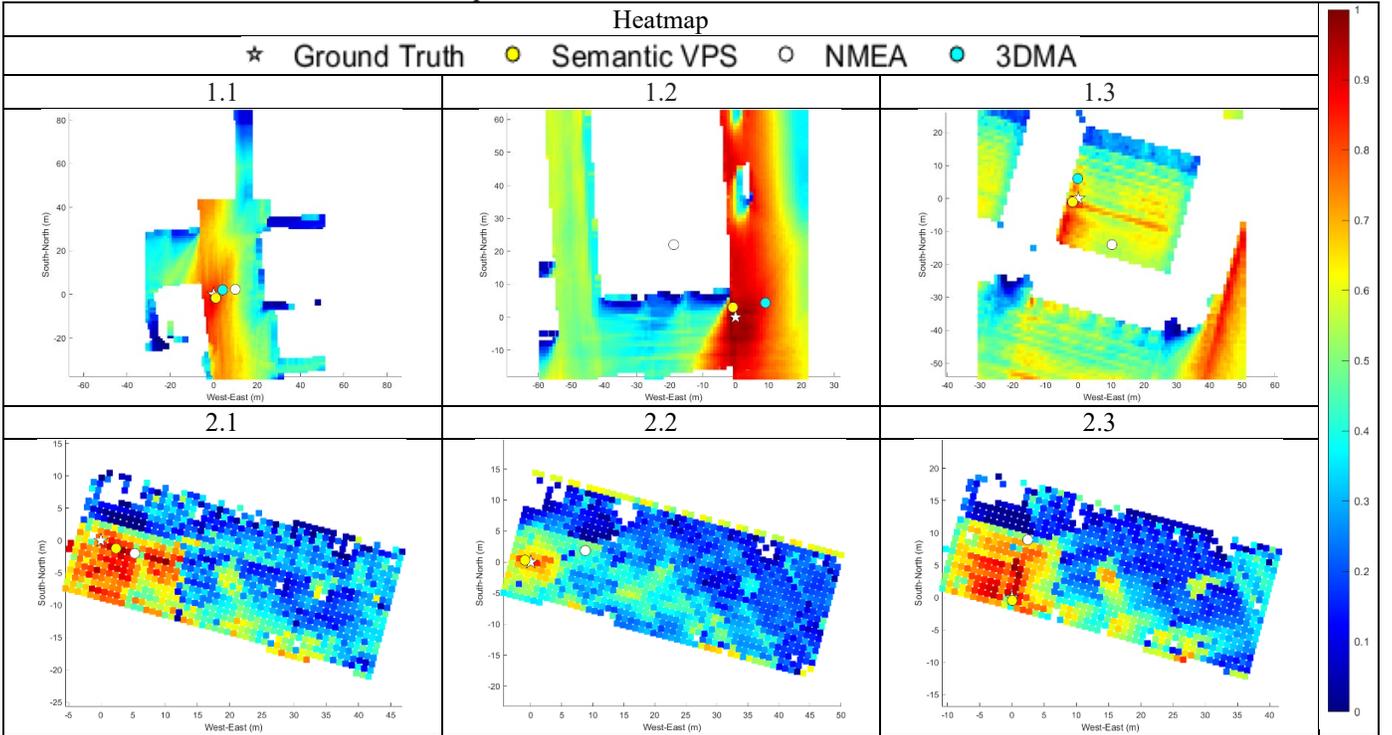

### 4.3 Semantic 3D map update results

The semantic 3D map update was analyzed based on the ideal smartphone image segmentation and compared to the chosen candidate image from the VPS solution. As shown in Table IV-IX, each class has their own respective RGB color: Stone (blue), Glass (green), Metal (orange), Banner (brown), Pedestrian (red), Chair (pink), Sky (black), Foliage (yellow) and Others (light blue). The calculation of the area of the material descriptor in **Loc. 1.1** is described in Section 4.4.

**Loc. 1.1** contains a large banner in the smartphone image that was not present in the semantic 3D map. The results show that the position of the updated banner deviates from the ground truth banner by 1.88m. It is important to note that the banner position deviation is similar in magnitude to the semantic VPS deviation of 1.96m. This phenomenon is expected as the position of the updated banner is estimated by the point cloud from the VPS image. Therefore, any positioning error in the semantic VPS will directly affect the positioning of the updated objects, the correlation of the errors is described in Section 4.4. The updated banner dimension is also 44% smaller in size compared to the ground truth banner. This can be explained by two errors. Firstly, it is likely that the shift in point cloud due to the VPS error will influence the dimension accuracy. However, since the banner is located on a flat building surface, ideally it will only increase the positioning error and not the dimensional error. Secondly, it is believed that the dimensional error is caused by the changed class detection error. As shown in the Table IV, the corners of the banner in the local change detection image have been slightly misestimated by the algorithm, producing a faulty area estimation of the changed class. **Loc. 1.2** has a significant material change from stone to glass on the building. This has proved too challenging to update on the semantic 3D map. Two requirements for successful material descriptor update are not fulfilled as mentioned in Sec. 3.9. Firstly, the material object in not captured entirely in the local change detector image. Therefore, it is unlikely to estimate an accurate dimension of the material as it may extend outside of the image. Secondly, the shape does not approximate a rectangle, which is not tested in this research. For these reasons, not updating the 3D semantic map will be the more appropriate choice. **Loc. 1.3** attempts to update a pedestrian in the semantic 3D map. The results show that the position of the updated pedestrian in the semantic 3D map deviates from the ground truth pedestrian by 1.57m. The two plausible positioning error of the updated pedestrian is due to the semantic VPS positioning error of 2.04m and changed class detection error. Although not meter level accuracy, the proposed solution demonstrates a suitable approach for dynamic 3D map update using user's smartphone images.

**Loc. 2.1** contains two chairs that are updated in the semantic 3D map. The results show that the semantic 3D map update solution can successfully identify multiple changed objects in the smartphone segmented image. The average position deviation of the chairs from the ground truth is 2.92m. The position deviation is similar to that of the semantic VPS deviation of 2.66m. The slight increase in deviation of the chair position compared to the VPS deviation is likely due to the error of the changed class detection. As shown in table IV, the chair's bottommost pixel in the image does not represent the position of the chair. In fact, the legs are spread across an area, none of which represents the centroid of the chair. As shown in the left chair, the algorithm successfully estimates the centroid of the chair, however it is difficult to extend the centroid to the ground to estimate the precise position of the chair. Therefore, a compromise is to use the width of the centroid's pixel, and the height of the chair's lowest pixel, to estimate the chair's position. **Loc. 2.2** demonstrates the effective removal of a chair in the semantic 3D map. The pixels of the entire chair in the local change detection are replaced with a material class. Therefore, the 3D position of the removed chair is used to search the closest chair descriptor in the semantic 3D map. The closest chair descriptor is then removed in the semantic 3D map. It is important to note this approach applies to all dynamic objects such as pedestrians as well. **Loc. 2.3** attempts to update a banner in the semantic 3D map. The experiment shows a very high semantic VPS positioning accuracy of 0.45m. However, as shown in table IV, the four-corner 3D points are not coplanar. It can be seen a slight error in VPS positioning can have a detrimental effect to the semantic 3D map update. Due to failing one of the requirements for successful material descriptor update, the banner cannot be updated.

TABLE IV. Map update of the classes detected in the smartphone image based on the proposed semantic 3D map update method.

| Loc. | Chosen Candidate Image with PC Overlay | Smartphone Segmented Image | Local Change Detection | Changed Class with PC Overlay | Updated Semantic 3D Map |
|---|---|---|---|---|---|
| 1.1 | | | | | |
| 1.2 | | | | | |
| 1.3 | | | | | |
| 1.4 | | | | | |
| 1.5 | | | | | |

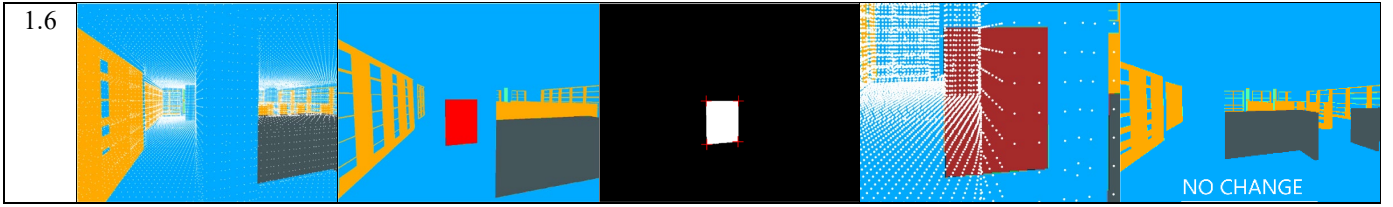

## 4.4 Accuracy evaluation of the semantic map update

We considered the two major types of error that will influence the semantic map update accuracy. We have tested in our experiments that VPS positioning error can contribute heavily to the semantic map update accuracy. Shown in **Loc. 2.3** where the slight shift in VPS positioning will lead to a large shift in the estimated point cloud of the class. We have also shown in **Loc. 1.1** that pixel detection error of the changed class will lead to a faulty position and area estimation of the changed class. Therefore, ideally, there should be no VPS positioning error and no pixel detection error. Getting these correct are very important, any shift in point cloud alignment can lead either no update to the semantic 3D map, or worse, incorrect update to the semantic 3D map. We further performed two experiments. Firstly, by updating a pedestrian located on a horizontal flat plane and increasing the VPS error to calculate the map update positioning accuracy. Secondly, by updating a material located on a vertical flat plane and increasing the VPS error to calculate the map update area accuracy. The pixel error is calculated based on the pixel difference between the estimated pixel of the changed class and the manual selection of the ground truth pixel. The positional error is calculated based on the 2D distance between the estimated descriptor position and the ground truth descriptor position. The area error is calculated based on the difference between the area of the estimated descriptor descripted in algorithm 3, and the area of the ground truth descriptor.

**Algorithm 3** Calculate area of material descriptor from four 3D coplanar points.

| | |
|---|---|
| **Input** | Detected four corners in world coordinate, $\mathbf{a}(x_1, y_1, z_1), \mathbf{b}(x_2, y_2, z_2), \mathbf{c}(x_3, y_3, z_3), \mathbf{d}(x_4, y_4, z_4)$ |
| **Output** | Area of material descriptor |

1  Calculate the area of $\Delta \mathbf{abc} = \frac{1}{2}\sqrt{\left|\overrightarrow{\mathbf{ab}}\right|^2 + |\overrightarrow{\mathbf{ac}}|^2 - (\overrightarrow{\mathbf{ab}} \cdot \overrightarrow{\mathbf{ac}})^2}$

2  Calculate the area of $\Delta \mathbf{acd}$

3  The total area of the material descriptor,
   $\Delta \mathbf{abcd} = \Delta \mathbf{abc} + \Delta \mathbf{acd}$

TABLE V. VPS positioning error and changed class pixel detection error on the semantic 3D map update method.

| Pedestrian on a horizontal flat plane | | Changed material on a vertical flat plane | |
|---|---|---|---|
| Candidate Image at Ground Truth with PC Overlay | Smartphone Segmented Image | Candidate Image at Ground Truth with PC Overlay | Smartphone Segmented Image |

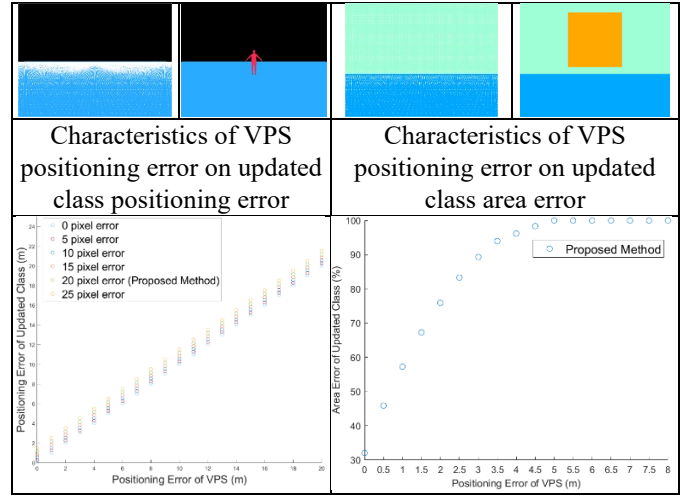

| Characteristics of VPS positioning error on updated class positioning error | Characteristics of VPS positioning error on updated class area error |
|---|---|

We purposely increased the VPS error from the ground truth by progressively moving the estimate VPS position in the opposite direction of the pedestrian and the changed material. The first result demonstrates that the proposed method has 20-pixel error. This has led to 1.17m of systematic positioning error of the updated class. It can also be seen that the positioning error of VPS will proportionally increase the positioning error of the update class. Therefore, for the semantic 3D map update to succeed in its updating purpose, it is detrimental to minimize the two errors. As shown in the second test, the area error of the update class is logarithmic to the VPS error. This is expected as the VPS error is multiplied to calculate the area of the updated class. It can be seen using the proposed method, there is already a 30% area error of the updated class due to the pixel error. This is followed by a deteriorating performance as VPS error increase, the map update algorithm fails to perform accurately, and once it reaches 5m of VPS error, the updated class will have 100% area error.

## V.  CONCLUSIONS AND FUTURE WORK

### 5.1 Conclusions

This paper proposes a novel semantic 3D map change detection and update by introducing semantics and virtual depth as a new source of information. In short, the changed classes are detected in the smartphone image and are then updated or removed in the semantic 3D map. Multiple computer vision algorithms were used to find the dimension and pose of the changed classes with robustness.

Existing 3D map update approaches using smartphone images updates only low-level features or requires the use of SfM for image depth estimation. This study is a method that extends existing research to formulate the 3D map update as a semantic problem using materials/objects as semantic information, and virtual LIDAR in the 3D map for image depth estimation. Our experiments demonstrate an alternate solution to updating 3D maps. The advantages for the semantic 3D map update solution are numerous:
- The formulation of change detection as a semantic problem enables us to apply the existing wide variety of advanced optimization/shape matching metrics to this problem.
- Materials and objects are diverse, distinctive, and distributed everywhere, hence the semantic information in an image is easy to recognize using artificial intelligence and deep learning.
- The utilization of virtual depth requires only one image and eliminates the need for SfM and depth estimation using multiple images.
- Identification and consideration of dynamic objects such that it can be used to update the 3D map for future dynamic 3D map applications.
- The use of descriptors makes it simple to add/remove/modify in the 3D map accurately.

Considering the results presented in this paper, we conclude the proposed method serves as an improved solution to existing advanced 3D map updating solutions using smartphone.

*5.2 Future Work*

Several potential future developments are suggested.
- Research has shown to identify a wide variety of materials and objects in images in the indoor environment. Therefore, it is suggested to develop and train a deep learning neural network to identify materials and objects in smartphone images in the outdoor and indoor environment for practical use [30]. Improvement in the deep learning neural network could also aid automatic segmentation of 3D maps, reducing the offline preparation time.
- The research has shown the VPS pose accuracy is proportionally corelated to the semantic 3D map update accuracy. Therefore, only the smartphone images with a reliable VPS pose should be used to update the 3D map. The use of a reliability algorithm to filter the reliable images and VPS poses will improve the semantic 3D map update accuracy.
- To add more common classes to differentiate means that given a large and high-quality dataset, the proposed method can be adapted to a variety of different uses.
- The virtual depth can also be used for AR during/after the 3D map update.
- It is suggested to use cloud-sourcing images to continuous update the 3D map from users.
- To maximize all available visual information, the combination of semantic VPS and feature VPS could yield better positioning performance.
- For dynamic positioning and map update, we can use a multiresolution framework, where the search starts from a big and sparse grid and is then successively refined on smaller and denser grids. Thus, the pose of the chosen candidate is used to refine a smaller search area.

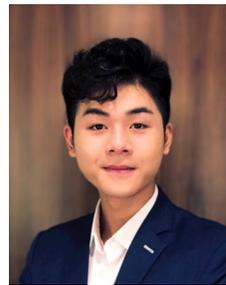
**Max Jwo Lem Lee** is currently a graduate of Bachelor of Engineering (Honours) in Aviation Engineering from the Hong Kong Polytechnic University. He has previously interned in Boeing as an engineer for 6 months and will be starting his PhD study in September 2021 with the Interdisciplinary Division of Aeronautical and Aviation Engineering, Hong Kong Polytechnic University. He has won the Hong Kong Techathon 2021. His other research interests include positioning in urban environments, indoor positioning, navigation, and unmanned aerial vehicle.

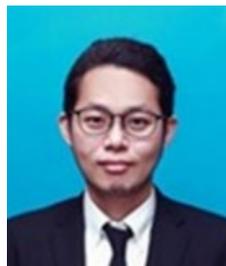
**Li-Ta Hsu (S'09-M'15)** received the B.S. and Ph.D. degrees in aeronautics and astronautics from National Cheng Kung University, Taiwan, in 2007 and 2013, respectively. He is currently an assistant professor with the Division of Aeronautical and Aviation Engineering, Hong Kong Polytechnic University, before he served as post-doctoral researcher in Institute of Industrial Science at University of Tokyo, Japan. In 2012, he was a visiting scholar in University College London, U.K. He is an Associate Fellow of RIN. His research interests include GNSS positioning in challenging environments and localization for pedestrian, autonomous driving vehicle and unmanned aerial vehicle.